\documentclass[10pt,journal]{IEEEtranTCOM}

\usepackage[english]{babel}
\usepackage[usenames]{color}
\usepackage[cp1250]{inputenc}
\usepackage{amsfonts}
\usepackage{amsthm}
\usepackage{graphicx}
\usepackage{epsfig}
\usepackage{mathrsfs}
\usepackage{amsmath}
\usepackage{algorithm}
\usepackage{algorithmic}
\usepackage{hyperref}

\theoremstyle{plain}

\begin{document}
\newcommand{\bea}{\begin{eqnarray}}
\newcommand{\eea}{\end{eqnarray}}
\newcommand{\be}{\begin{equation}}
\newcommand{\ee}{\end{equation}}
\newcommand{\beas}{\begin{eqnarray*}}
\newcommand{\eeas}{\end{eqnarray*}}
\newcommand{\bs}{\backslash}
\newcommand{\bc}{\begin{center}}
\newcommand{\ec}{\end{center}}
\def\SC {\mathscr{C}}

\title{Improving SGD convergence \\ by online linear regression of gradients \\ in multiple statistically relevant directions}
\author{\IEEEauthorblockN{Jarek Duda}\\
\IEEEauthorblockA{Jagiellonian University,
Golebia 24, 31-007 Krakow, Poland,
Email: \emph{dudajar@gmail.com}}}
\maketitle

\begin{abstract}
Deep neural networks are usually trained with stochastic gradient descent (SGD), which minimizes objective function using very rough approximations of gradient, only averaging to the real gradient. Standard approaches like momentum or ADAM only consider a single direction, and do not try to model distance from extremum - neglecting valuable information from calculated sequence of  gradients, often stagnating in some suboptimal plateau.
Second order methods could exploit these missed opportunities, however, beside suffering from very large cost and numerical instabilities, many of them attract to suboptimal points like saddles due to negligence of signs of curvatures (as eigenvalues of Hessian).

Saddle-free Newton method (SFN)~\cite{SFN} is a rare example of addressing  this issue - changes saddle attraction into repulsion, and was shown to provide essential improvement for final value this way. However, it neglects noise while modelling second order behavior, focuses on Krylov subspace for numerical reasons, and requires costly eigendecomposion.

Maintaining SFN advantages, there are proposed inexpensive ways for exploiting these opportunities. Second order behavior is linear dependence of first derivative - we can optimally estimate it from sequence of noisy gradients with least square linear regression, in online setting here: with weakening weights of old gradients. Statistically relevant subspace is suggested by PCA of recent noisy gradients - in online setting it can be made by slowly rotating considered directions toward new gradients, gradually replacing old directions with recent statistically relevant. Eigendecomposition can be also performed online: with regularly performed step of QR method to maintain diagonal Hessian. Outside the second order modeled subspace we can simultaneously perform gradient descent.
\end{abstract}
\section{Introduction}
In many machine learning situations we search for parameters $\theta\in \mathbb{R}^D$ (often in millions) minimizing some \emph{objective function} $F:\mathbb{R}^D\to\mathbb{R}$ averaged over given (size $n$) dataset:
\be F(\theta) =\frac{1}{n}\sum_{i=1}^n F_i(\theta) \ee
where $F_i(\theta)$ is objective function for $i$-th object of dataset. We would like to minimize $F$ through gradient descent, however, dataset can be very large, and calculation of $\nabla_\theta F_i$ is often relatively costly, for example through backpropagation of neural network. Therefore, we would like to work on gradients calculated from succeeding subsets of dataset (mini-batches), or even single objects (original SGD). Additionally, objective function can have some intrinsic randomness.

In stochastic gradient descent (SGD) framework~\cite{SGD}, we can ask for some approximation of gradient for a chosen point $\theta^t$ in consecutive times $t$:
\be g^t=\nabla_\theta  F^t(\theta^t)\qquad \qquad (\approx \nabla_\theta F(\theta^t))\ee
where $F^t$ corresponds to average over succeeding subset (mini-batch) or even a single element $(F^t=F_i)$ - we can assume that $g^t$ averages to the real gradient over time. In this philosophy, instead of using the entire dataset in a single point of the space of parameters, we do it in online setting: improve the point of question on the way to use more accurate local information.

To handle resulting noise we need to extract statistical trends from $g^t$ sequence while optimizing evolution of $\theta^t$, providing fast convergence to local minimum, avoiding saddles and plateaus. As calculation of $g^t$ is relatively costly and such convergence often stagnates in a plateau, improving it with more accurate (and costly) modelling might provide significant benefits especially for deep learning applications - very difficult to train efficiently due to long reason-result relation chains.

A natural basic approach is working on momentum~\cite{momentum}: evolve $\theta$ toward some (e.g. exponential moving) average of $g^t$. There are also many other approaches, for example popular AdaGrad~\cite{adagrad} and ADAM~\cite{adam} estimating both average gradient and coordinate-wise squared gradient to strengthen underrepresented coordinates.\\

Such standard simple methods leave two basic opportunities for improved exploitation of statistical trends from $g^t$ sequence, discarding valuable information this sequence contains. We would like to practically use them here:
\begin{enumerate}
  \item They do not try to estimate distance from local extremum ($\nabla F=0$), which is suggested in linear behavior of gradients, and could allow to optimize the crucial choice of step size: which should be increased in plateaus, decreased to avoid jumping over valleys. It can be exploited in second order methods, however, directly calculating inverse Hessian from noisy gradients is numerically problematic. Instead, we will extract it from statistics of noisy gradients in an optimal way: estimate this linear behavior of gradients by least squares linear regression, reducing weights of old gradients. Controlling sign of curvatures we can handle saddles this way - correspondingly attracting ($\lambda<0$) or repelling ($\lambda<0$).
  \item First order methods focus on information in only a single considered direction, discarding the remaining. Modelling based on information in multiple directions allows for optimization step in all of them simultaneously, for example attracting in some and repelling in others to efficiently pass saddles. As full Hessian is often too costly, a natural compromise is second order modelling in $d$ dimensional linear subspace, usually $d \ll D$. One difficulty is locally choosing subspace where the action is, for example as largest eigenvalues subspace of PCA of recent noisy gradients. It can be turned into online update of considered directions by slowly rotating them toward new noisy gradients, extracting their statistical trends. While not necessary, maintaining nearly diagonal Hessian in the considered subspace should improve performance - it can be made e.g. by making linear regression in entire subspace and periodically performing diagonalization, or regularly QR method~\cite{QR} step.
\end{enumerate}

Both these opportunities could be exploited if trying to model full Hessian $H$, but it is usually much too costly: requiring at least $O(D^2)$ memory and regular computations. However, we can focus on some chosen number $d\leq D$ of looking most promising directions (as modelled eigenvectors), reducing this memory and computational cost to $\approx O(dD)$.

Many second order methods neglect signs of curvatures, e.g. natural gradient~\cite{natural}: $\theta\leftarrow \theta-H^{-1}g$ attracting to near point with zero gradient, which is usually a saddle. It also concerns quasi-Newton methods like L-BFGS~\cite{L-BFGS}. Many other methods approximate Hessian with positive-defined, trying to pretend that non-convex function is locally convex, again attracting not only to local minima - for example using Gauss-Newton, Levenberg-Marquardt, or analogous Fisher information matrix (e.g. K-FAC~\cite{K-FAC}). Negative curvature is also neglected in covariance matrix based methods like TONGA~\cite{TONGA}. Nonlinear conjugated gradient method assumes convectiveness.

However, saddles are believed to be very problematic in such training. While there is belief that nearly all local minima are practically equally good, they are completely dominated by saddles - which number is exponentially (with dimension) larger than of minima. Additionally, plateaus are often formed near them - requiring much larger steps to be efficiently passed. Hence it is beneficial to not only use second order methods, but those which can handle saddles - which include signs of curvatures into considerations, instead of just ignoring it.\\

Very rare example of such methods is saddle-free Newton (SFN)~\cite{SFN}, changing step sign for negative curvature directions: from attraction to saddle into repulsion. It has shown to lead to significantly better parameters, literally a few times lower error rate on MNIST  while compared with other methods: stagnating in suboptimal solutions with strong negative curvatures (saddles, plateaus). SFN was able to weaken these negative eignevalues a few orders of magnitude. The importance of not neglecting negative curvature is also presented e.g. in \cite{hess}, showing that such rare negative curvature directions allow for significant improvements of value.

However, SFN leaves some improvement opportunities, we would like to exploit here:
\begin{enumerate}
  \item SFN tries to directly calculate second order behavior, neglecting stochastic nature of base of this calculation, what results in numerical instabilities for required Hessian inversion. To extract it from statistics instead, we can see second order behavior as linear trend in the first order behavior, which can be found in an optimal way by linear regression of the noisy gradients, for example with weakening weights of the older ones.
  \item While we need to restrict modelling to a subspace, SFN uses Krylov subspace for convenience of numerical procedure. Instead, recent noisy gradients suggest locally interesting directions, which can be extracted through PCA or its online analogue discussed here.
  \item Instead of calculating eigendecomposition in a given point, we can do it in online setting: split into regularly performed steps maintaining diagonal Hessian, like of QR method.
  \item As we can practically use second order model only for low dimensional subspace, in the remaining directions we can simultaneously perform gradient descent.
\end{enumerate}

This paper is work in progress, requiring experimental investigation for choosing the details. The v4 version is completely rewritten - previous were based on gradient agreement, what is problematic for stability, only 1D linear regression was suggested. This version is completely based on gradient linear regression, including multidimensional.
\section{1D case with linear regression of derivatives}
Estimating second order behavior from noisy gradients is a challenging task, especially if we would like to distinguish signs of curvatures - what is required to avoid saddle attraction. Close to zero Hessian eigenvalues can change sing due to this randomness, changing predicted zero gradient position from one infinity to another.

Hence this estimation needs to be performed in a very careful way, preferably extracting statistics from a large number of such noisy gradients. What we are in fact interested in is position of zero derivative point in each considered direction, determined by liner trend of first derivative. An optimal way to extract linear trend is least squares linear regression - applied to sequence of gradients here, preferably with reduced weights of old noisy gradients to include two approximations: that function is only locally modeled as second degree polynomial, and that we would like to slowly rotate the considered directions.

Let us focus on 1D case in this Section, then we will do it in multiple directions (e.g. a few from a million) - separately in each direction for diagonalized, or combined to maintain nearly diagonal Hessian. The choice of such subspace will explore recent statistically relevant directions, in the remaining we can perform casual gradient descent.
\subsection{1D static case - parabola approximation}
Let us start with 1D case, with parabola model first:
$$f(\theta)=h+\frac{1}{2}\lambda(\theta-p)^2\qquad\qquad f'(\theta)=\lambda (\theta-p) $$
and MSE optimizing its parameters for $(\theta^t,g^t)$ sequence:
$$\arg\min_{\lambda,p}\, \sum_t w^t (g^t-\lambda(\theta^t-p))^2\quad\textrm{for some weights } (w^t)$$
For parabola and $t=1,\ldots,T$ times we can choose uniform weights $w^t=1/T$. Later we will use exponential moving average - reducing weights of old noisy gradients, seeing such parabola as only local approximation. The necessary $\partial_p=\partial_\lambda=0$ condition (neglecting $\lambda=0$ case) becomes:
$$ \sum_t w^t (g^t-\lambda(\theta^t-p)) = 0 =\sum_{t}w^t (\theta^t-p)(g^t -\lambda \theta^t +\lambda p) $$
$$\overline{g}-\lambda\overline{\theta}+s\lambda p = 0 = \overline{\theta g}-\lambda\overline{\theta^2}+2\lambda p\overline{\theta}-p\overline{g}-s\lambda p^2$$
for averages:
$$s=\sum_t w^t\qquad\overline{\theta}=\sum_t w^t \theta^t\qquad  \overline{g}=\sum_t w^t g^t$$
\be  \overline{\theta g}=\sum_t w^t \theta^t g^t\qquad\quad \overline{\theta^2}=\sum_t w^t (\theta^t)^2 \label{fa}\ee
Their solution is (least squares linear regression):
$$ \lambda=\frac{s\,\overline{g\theta}-\overline{g}\,\overline{\theta}}
{s\,\overline{\theta^2}-\overline{\theta}^2 }\qquad \textrm{clipped e.g.:}\quad \lambda=\frac{c(s\,\overline{g\theta}-\overline{g}\,\overline{\theta})}
{s\,\overline{\theta^2}-\overline{\theta}^2 }$$
\be p=\frac{\lambda\overline{\theta}-\overline{g}}{s\lambda}=\frac{\overline{\theta^2}\,\overline{g} - \overline{\theta}\,\overline{g\theta }}
{\overline{\theta}\,\overline{g}-s\,\overline{\theta g}}\label{reg}\ee
Where some "clipping" is required to avoid $\lambda\approx 0$, e.g. for some minimal value $\epsilon>0$,  $c(x)=\textrm{sign}(x) \min(|x|,\epsilon)$.

Observe that $\lambda$ estimator is $(g,\theta)$ covariance divided by variance of $\theta$ (positive if not all equal).
\subsection{1D online update by exponential moving average}
Objective function is rather not exactly parabola - should be only locally approximated this way. It can be handled by increasing recent weights in the above averages, weakening influence of the old noisy gradients. Its simplest realization is through exponential moving averages $w^t\propto \beta^{-t}$ for $\beta\in(0,1)$ usually $\beta\in(0.9,0.999)$, allowing to inexpensively update such averages for a given moment, starting with 0 in $t=0$:
$$\overline{\theta}^t=\beta\, \overline{\theta}^{t-1}+(1-\beta)\,\theta^t=(1-\beta)\sum_{t'=1}^t \beta^{t-t'}\, \theta^{t'}$$
$$\overline{g}^t=\beta\, \overline{g}^{t-1}+(1-\beta)\,g^t$$
$$\overline{\theta g}^t=\beta\, \overline{\theta g}^{t-1}+(1-\beta)\,\theta^t g^t$$
$$\overline{\theta^2}^t=\beta\, \overline{\theta^2}^{t-1}+(1-\beta)\,(\theta^t)^2$$
\be s^t=(1-\beta)\sum_{t'=1}^t \beta^{t-t'}=1-\beta^t \label{avg} \ee
The $s^t$ is analogous e.g. to bias in ADAM, in later training it can be assumed as just $s=1$.

It might be worth modifying $\beta$, e.g. increasing it for larger step to reduce weights of far gradients.
\subsection{1D linear regression-based SGD method}
Linear regression requires values in at least two points, hence there is needed at least one step (better a few) of a different method to start using linear regression, for example just SGD - going toward stochastic gradients, updating averages (\ref{avg}) starting from initial $\overline{\theta}^0=\overline{g}^0=\overline{\theta g}^0=\overline{\theta^2}^0=s^0=0$.

Then we can start using linear model for derivative: $f'(\theta)\approx \lambda (\theta-p)$, using updated parameters from (\ref{reg}) regression.

Getting $\lambda>0$ curvature, the modeled optimal position would be $\theta=p$. However, as we do not have complete confidence in this models, and would like to work in online setting, a safer step is $\theta\leftarrow \theta+\alpha (p-\theta)$ for $\alpha\in(0,1]$ parameter describing trust in the model, which generally can vary e.g. depending on estimated uncertainty of parameters. Natural gradient method corresponds to $\alpha=1$ complete trust.

Getting $\lambda<0$, minimizing modelled parabola would take us to infinity, hence we need some arbitrary choice for these negative curvature directions. Saddle-free Newton kind of chooses $-\alpha$ in these directions, experiments in $\cite{hess}$ suggest to use $\sim 0.1$ of gradient projection for such directions.

The $\lambda \approx 0$ case can correspond to plateau, or to inflection point: switching between convex and concave behavior. These are very different situations: in the former we should maintain larger step size for a longer time, in the latter we need to be more careful as $\lambda=0$ would correspond to $p$ in infinity. These two cases could be distinguished for example by looking at evolution of $\lambda$ (higher order method), or just at its local variance: reduce step if it is large.

While it leaves opportunities for improvements, for simplicity we can for example use SFN-like step: just switching sign for $\lambda<0$ directions. Applied clipping prevents $\lambda\approx 0$ cases, alternatively we could e.g. replace sign with $\tanh$:

\be \theta^{t+1}=\theta^t+\alpha\,\textrm{sign}(\lambda^t)\, (p^t-\theta^t) \label{step}\ee
\section{Multidimensional case}
We can now take it higher dimensions, what will be done in 3 steps here: first directly for the entire space, then with periodic or online diagonalization updating basis rotation, and finally as a linear subspace of a higher dimensional space - additionally rotated toward new statistically relevant directions. The next Section contains algorithm for such final method for improving convergence of SGD.
\subsection{Direct multivariate approach}\label{hesder}
Second degree polynomial parametrization in $d$ dimensional space analogously becomes:
$$f(\theta)=h+\frac{1}{2}(\theta-p)^T H(\theta-p)\qquad\qquad \nabla f=H(\theta-p)$$
For Hessian $H\in\mathbb{R}^{d\times d}$ and $p\in \mathbb{R}^d$ position of saddle or extremum. Least square linear regression would like to analogously minimize:
$$\arg\min_{H,p}\ \sum_{i,t} w^t \left(g^t_i-\sum_k H_{ik}(\theta^t_k-p_k)\right)^2 $$
Getting analogous necessary conditions. First for $\partial_{p_j}=0$ (neglecting generic case of getting to kernel of $H$):
$$\forall_{j}\quad \sum_{t,i} w^t \left(g^t_i-\sum_k H_{ik}(\theta^t_k-p_k)\right)H_{ij}=0 $$
$$ \forall_i\qquad \overline{g_i} - \sum_k H_{ik}\overline{\theta_k}+s\sum_k H_{ik}p_k=0 $$
\be \overline{g}=H\overline{\theta}-s\,Hp=H(\overline{\theta}-s\,p) \label{eq1}\ee
for $\overline{g}_i=\overline{g_i}$, $\overline{\theta}_i=\overline{\theta_i}$ vectors of averaged values as previously, $s=\sum_t w^t$. For $\partial_{H_{ij}}=0$ we get:
$$\forall_{i,j}\quad \sum_t w^t (\theta^t_j-p_j)\left(g^t_i-\sum_k H_{ik}(\theta^t_k-p_k)\right)=0 $$
$$\overline{g_i\theta_j}-\overline{g_i}p_j=\sum_k H_{ik} \left(\overline{\theta_k\theta_j}-p_k\overline{\theta_j}-
\overline{\theta_k}p_j+s\,p_k p_j\right)$$
$$\overline{g\theta}-\overline{g}p^T=H\overline{\theta\theta}-Hp\overline{\theta}^T
-H(\overline{\theta}-sp)p^T$$
where the last is matrix equation with $\overline{g\theta}_{ij}=\overline{g_i\theta_j}$,  $\overline{\theta\theta}_{ij}=\overline{\theta_i\theta_j}$ averages. Substituting (\ref{eq1}) twice $(Hp=s^{-1}(H\overline{\theta}-\overline{g}))$ we get:
$$\overline{g\theta}\overset{(\ref{eq1})}{=} H\overline{\theta\theta}-Hp\overline{\theta}^T\overset{(\ref{eq1})}{=}
H\overline{\theta\theta}-s^{-1}(H\overline{\theta}-\overline{g})\overline{\theta}^T$$
$$s\overline{g\theta}=
sH\overline{\theta\theta}-H\overline{\theta}\overline{\theta}^T+\overline{g}\overline{\theta}^T$$
\be H=\left(s\overline{g\theta}-\overline{g}\,\overline{\theta}^T\right)
\left(s\overline{\theta\theta}-\overline{\theta}\,\overline{\theta}^T\right)^{-1}
\label{regm}\ee
analogous to $\lambda$ formula (\ref{reg}) in 1D, replacing covariance with covariance matrices, denominator is positive defined. Using
\be p=(\overline{\theta}- H^{-1}\overline{g})/s \label{regmp} \ee
we get the $\nabla f=0$ position: extremum or saddle of degree 2 polynomial modelling our function.\\

To treat it only as local model, in online setting we can analogously calculate averages as exponential moving average, e.g.
$$\overline{\theta_i \theta_j}^{t+1}=\beta\,\overline{\theta_i \theta_j}^{t}+(1-\beta)\,\theta_i^t \theta_j^t$$

Now SFN-like approach would be calculating eigendecomposition $H=O^T \Lambda O$, then performing
$$\theta^{t+1}=\theta^t+O^T \textrm{sign}(\Lambda)\, O\, (\theta^t-p^t)$$
Where $\textrm{sign}(\Lambda)$ means applying sign to each coordinate of this diagonal matrix - it turns attraction into repulsion for negative curvature directions to handle saddles. There is also needed some clipping to handle $\lambda\approx 0$.

To avoid costly diagonalization in every step, for online method let us now discuss doing it periodically.

\subsection{Periodic diagonalization - currently main approach}
Diagonalization is relatively costly and using inaccurate one seems not a problem, e.g. completely omitting it is analogous to using gradient method instead of conjugated gradients.

Hence we can perform it periodically - every some number of steps, modifying the considered basis: in which we can assume that Hessian is nearly diagonal.

Let $(v^t_i)_{i=1..d}$ be basis in moment $t$, which is approximately orthonormal (no need for perfect): $v_i^t\cdot v_j^t\approx \delta_{ij}$. Denote:
\be \theta^t_{\cdot i}=\theta^t\cdot v^t_i\qquad\qquad  g^t_{\cdot i}=g^t\cdot\, v^t_i\ee
as coordinates in current basis. We can update multidimensional averages for linear regression in these coordinates, e.g.
$$\overline{\theta_i \theta_j}^{t+1}=\beta\,\overline{\theta_i \theta_j}^{t}+(1-\beta)\,\theta_{\cdot i}^t\, \theta_{\cdot j}^t$$
Then update 1D regression (\ref{reg}) independently for each coordinate $i=1,\ldots,d$, using diagonals: $\overline{g_i \theta_i}$, $\overline{\theta_i \theta_i}$, and perform step (\ref{step}) in each coordinate (with reduced trust $\alpha$). The next Section contains complete algorithm.

Additionally, while in most steps the basis is unchanged,
every some number of steps we should improve digitalization: estimate Hessian from averages using (\ref{regm}), find its eigendecomposition $H=O^T \Lambda O$, use it to rotate the basis $[v_1,\ldots,v_d]^T \leftarrow O[v_1,\ldots,v_d]^T$ (matrix with $v_i$ as columns) and calculated averages:
$$\overline{\theta}\leftarrow O \overline{\theta}\qquad
\overline{g}\leftarrow O \overline{g}\qquad\overline{\theta\theta}\leftarrow O\overline{\theta\theta}O^T\qquad\overline{g\theta}\leftarrow O\overline{g\theta}O^T$$

\subsection{Online diagonalization }
There is also a possibility to perform diagonalization in online setting - split it into regular less expensive steps maintaining nearly diagonal Hessian.

Looking at Hessian formula (\ref{regm}), we can start from diagonal $H$ and ask for let say first order correction during step of updating the averages - introducing tiny nondiagonal terms.

Then we can perform step for example of QR algorithm~\cite{QR}: decomposing matrix $A=QR$ and multiplying in opposite order: $B=RQ=Q^T QRQ=Q^T AQ$, this way reducing nondiagonal terms. Its cost can be reduced if neglecting products of non-diagonal terms, also for rotating the basis with such $Q$ close to identity matrix.

To avoid directly calculating Hessian with (\ref{regm}), we could try to separately evolve $\left(s\overline{g\theta}-\overline{g}\,\overline{\theta}^T\right)$ and
$\left(s\overline{\theta\theta}-\overline{\theta}\,\overline{\theta}^T\right)^{-1}$,
using $(AB^{-1})'=A'B^{-1}-AB^{-1}B'B^{-1}$ formula and first order steps for updating the averages.

However, it seems still relatively costly, would need to recalculate Hessian sometimes due to inaccuracies, brings additional complications - the details of possible improvements it can bring are left for further work.

\subsection{Modeling subspace for very high dimensions}
As the original space of parameters has often huge dimension ($D$) in machine learning applications, often in millions, for practical optimization we would like to model Hessian only for some of them ($d\ll D$), e.g. a few. In the remaining we can simultaneously e.g. perform gradient descent.

Modelling just $d=2$ directions, in contrast to $d=1$, has a chance to see both attracting and repelling direction near a saddle to efficiently handle them. As negative eigenvalues have often lower absolute values, what we can see e.g. in \cite{hess}, there is rather required larger $d$ to include some negative curvature directions in the model, like $d=10$.

To work in a linear subspace, analogously as for periodic online diagonalization, we can consider evolving $(v^t_i)_{i=1..d}$ basis, this time with vectors from the large space: $v^t_i\in \mathbb{R}^D$.

The question is how to choose these $d$ directions in much larger $D$ dimensional space? We would like to find where the action locally is, what is suggested by directions of the fastest change: (noisy) gradients. To extract multiple relevant directions from their statistics, a natural way is performing PCA and taking subspace spanned by eigenvectors corresponding to the largest $d$ absolute eigenvalues - getting $d$ dimensional subspace with the lowest average Euclidean distance from projections. However, PCA would require construction and diagonalization of huge $D\times D$ covariance matrix.

Hence we would like to use only $(v^t_i)_{i=1..d}$ basis as the current description, and modify it accordingly to part of gradient it cannot represent:
\be \tilde{g}^t:=g^t-\sum_{i=1}^d (g^t\cdot v_i^t)\,v_i^t \ee
which can be also directly used for simultaneous gradient descent. A simple way to update the basis is just adding $v_i\ +=\gamma_i\,\tilde{g}^t$ to each vector. This way recent statistically relevant directions would gradually become represented by the used basis, replacing locally insignificant directions.

To maintain $v_i\cdot v_j\approx\delta_{ij}$, sometimes improved with orthonormalization step, let us assume it in time $t$ and find matrix $\Gamma\approx I$ to satisfy it in $t+1$ in below form:
\be v^{t+1}_i=\Gamma_{ii}\,v^t_i+\sum_{j\neq i} \Gamma_{ij} v^t_j+\gamma_i\, \tilde{g}\label{explore}\ee
$$1\approx v^{t+1}_i\cdot v^{t+1}_i\approx\Gamma_{ii}^2+\sum_{j\neq i} (\Gamma_{ij})^2 +
\gamma_i^2\,\|\tilde{g}\|^2$$
$$i\neq j\,: \qquad 0\approx v^{t+1}_i\cdot v^{t+1}_j \approx\Gamma_{ii}\Gamma_{ji}+\Gamma_{ij}\Gamma_{jj}+\gamma_i\gamma_j\|\tilde{g}\|^2$$
Neglecting higher order terms (e.g. $\Gamma_{ii}\Gamma_{ji}\approx \Gamma_{ji}$), and taking symmetric $\Gamma_{ij}=\Gamma_{ji}$, we can choose $(i\neq j)$:
$$\Gamma_{ij}=-\frac{1}{2} \gamma_i\gamma_j\|\tilde{g}\|^2\qquad \qquad
\Gamma_{ii}=1-\frac{1}{2}\gamma_i^2\|\tilde{g}\|^2$$

Further pseudocode simplifies it for $\gamma_i=\gamma$ choice - rotating all basis vectors with the same strength. It might be also worth to consider e.g. being more conservative for large $|\lambda|$ directions - try to mostly rotate those of low $|\lambda|$, what can be obtained e.g. by using $\gamma_i\propto |\lambda_i|^{-\kappa}$ e.g. for $\kappa=1/2$.

We slowly loose orthonormality this way, hence orthonormalization should be performed from time to time, e.g. every some number of steps, or if not passing some test of orthonormality. While Gram-Schmidt depends on vector order, we can e.g. use approximate but symmetric orthonormalization step:
\be \forall_i\ u_i \leftarrow v_i-\sum_{j\neq i} (v_i\cdot v_j)\,v_j\quad\textrm{then}\quad\forall_i\ v_i \leftarrow \frac{u_i}{\|u_i\|_2}\ee

As we neglect all but $d$ direction, we can additionally make gradient descent $\theta\leftarrow \theta+\eta\tilde{g}$ for a standard choice of $\eta$.
\section{Algorithm example}
This Section summarizes a basic choice of algorithm using periodic diagonalization. For simplicity it neglects time index.

\textbf{Initialization} - choose:
\begin{itemize}
\item $d$ number of considered directions (could vary),
\item $\alpha\in (0,1]$ describing confidence in model, step size - can be increased in the beginning,
\item $\beta\in (0,1)$ constant in exponential moving average, weight of old gradients drops $\propto \beta^{\Delta t}$. Generally can depend e.g. on step size - be increased for larger steps.
\item tiny $\gamma>0$  describing speed of exploration of new directions, can be e.g. increased in the beginning,
\item tiny $\epsilon>0$ for clipping - handling $\lambda\approx 0$ situations,
\item optional $\eta>0$ for neglected directions gradient descent,
\item $\theta$ - initial parameters, e.g. chosen randomly using probability distribution of parameters of given type of network,
\item $s=0\in\mathbb{R}$, $\overline{g}=\overline{\theta}=0\in\mathbb{R}^d$, $\overline{g\theta}=\overline{\theta\theta}=0\in\mathbb{R}^{d\times d}$
\item $(v_1,\ldots,v_d)$ initial basis - for example as random size $d$ orthonormal set of vectors $v_i\in \mathbb{R}^D$.
\end{itemize}

\textbf{Initial model training} - perform some number of steps of a different method like SGD $(\theta\leftarrow \theta-\eta g)$, in each step updating averages (2 below) and basis (5 below, $\gamma$ can be increased). Finally diagonalize Hessian (6 below) and orthonormalize the basis (7 below).

\textbf{Until some convergence condition} do \textbf{optimization step}:
\begin{enumerate}
  \item \textbf{Calculate stochastic gradient} $g\leftarrow g^t$ in current postion $\theta^t=\theta$ for current minibatch,
  \item \textbf{Update averages} for linear regression in $(v_i)$ basis:
      $$\forall_{i=1..d}\quad\theta_{\cdot i}\leftarrow \theta\cdot v_i\qquad g_{\cdot i}\leftarrow g\cdot v_i$$
      $$\forall_{i=1..d}\quad \overline{\theta}_i\leftarrow \beta\, \overline{\theta}_i+(1-\beta)\,\theta_{\cdot i}$$
      $$\forall_{i=1..d}\quad \overline{g}_i\leftarrow \beta \, \overline{g}_i+(1-\beta)\,g_{\cdot i}$$
      $$\forall_{i,j=1..d}\quad \overline{\theta\theta}_{ij}\leftarrow \beta\, \overline{\theta\theta}_{ij}+(1-\beta)\,\theta_{\cdot i}\,\theta_{\cdot j}$$
      $$\forall_{i,j=1..d}\quad \overline{g \theta}_{ij}\leftarrow \beta\, \overline{g\theta}_{ij}+(1-\beta)\,g_{\cdot i}\,\theta_{\cdot j}$$
      $$s\leftarrow \beta s+(1-\beta) $$
  \item \textbf{Calculate curvatures and positions} assuming diagonal Hessian:
$$\forall_{i=1..d}\qquad \lambda_i\leftarrow\frac{c(s\,\overline{g\theta}_{ii}-\overline{g}_i\,\overline{\theta}_i)}
{s\,\overline{\theta\theta}_{ii}-(\overline{\theta}_i)^2}\qquad \quad p_i\leftarrow \frac{\lambda_i\overline{\theta}_i-\overline{g}_i}{s\lambda_i}$$
with example of clipping: $c(x)=\textrm{sign}(x)\,\min(|x|,\epsilon)$.
  \item Perform \textbf{proper step} for position, for example:
  $$\theta\leftarrow\theta+\alpha\sum_{i=1}^d\, \textrm{sign}(\lambda_i)\, (p_i-\theta_{\cdot i})\,v_i$$
\item \textbf{Explore new directions} - outside current subspace:
$$\tilde{g}:=g-\sum_{i=1}^d (g\cdot v_i)v_i\qquad\qquad \bar{v}=\frac{1}{2}\gamma^2\, \|\tilde{g}\|^2 \sum_{i=1}^d v_i$$
$$\forall_{i=1..d}\qquad  v_i\leftarrow v_i+\gamma\, \tilde{g}-\bar{v}$$
$$\textrm{optionally do gradient descent:}\quad \theta\leftarrow \theta-\eta \tilde{g} $$
\item Every some number of steps \textbf{diagonalize Hessian}:
$$H\leftarrow\left(s\overline{g\theta}-\overline{g}\,\overline{\theta}^T\right)
\left(s\overline{\theta\theta}-\overline{\theta}\,\overline{\theta}^T\right)^{-1}$$
$$\textrm{diagonalize:}\qquad H=O^T \Lambda O$$
$${\theta}\leftarrow O \overline{\theta}\qquad \overline{g}\leftarrow O \overline{g}$$
$$\overline{\theta\theta}\leftarrow O\overline{\theta\theta}O^T \qquad\overline{g\theta}\leftarrow O\overline{g\theta}O^T$$
$$\forall_i\ u_i\leftarrow \sum_j O_{ij}v_j\qquad\textrm{then}\qquad \forall_i\  v_i\leftarrow u_i$$
\item Every some number of steps \textbf{improve orthonormality}:
$$\forall_{i=1..d}\quad u_i \leftarrow v_i-\sum_{j\neq i} (v_i\cdot v_{j})v_{j}\qquad v_i \leftarrow \frac{u_i}{\|u_i\|_2} $$
\end{enumerate}
\section{Conclusions and further work}
There were presented general ideas for second order optimization methods:
\begin{itemize}
\item estimating second order behavior from linear trend of (noisy) gradients - using least square linear regression,
\item focused on online setting: evolution split into regular inexpensive steps exploiting local behavior,
\item using inexpensive adaptation of subspace to statistically relevant directions in sequence of gradient,
\item optimized also for non-convex situation, handling saddles by considering signs of curvatures,
\item hybrid with first order model - simultaneously using stochastic gradient outside second order model subspace.
\end{itemize}
While they can be useful also for other situations, the main purpose here is optimizing SGD - algorithm for such application is finally suggested. Choosing the details like parameters or minibatch size will require further experimental work. There are also many questions and opportunities to explore, for example:
\begin{itemize}
 \item Beside the question of choosing parameters including minibatch size, it might be worth to evolve them - e.g. increase steps $\alpha$ in the beginning, lower $\beta$ for larger steps for faster forgetting of far gradients, increase $\gamma$ in the beginning for faster search of relevant directions.
 \item While for $\lambda>0$ we should just go toward minimum of modeled parabola, the $\lambda<0$ case needs some arbitrary choice of step size. As in SFN we just switch sign here, however, it seems unlikely that it is the optimal way, there is flexibility to customize it.
  \item Improving way of handling $\lambda\approx 0$ situations, e.g. various ways for clipping, maybe using higher order behavior to distinguish inflection point from plateau.
  \item Online diagonalization might offer improvements.
  \item It might be worth weakening external basis rotation for higher absolute eigenvalues e.g. $\gamma_i=|\lambda_i|^{-\kappa}$, the question is how much: e.g. what power to choose.
  \item It might be worth adding estimation of uncertainty especially of positions $p$, and modify step size $\alpha$ accordingly.
  \item Subspace dimension $d$ might evolve depending on local situation, e.g. by removing low curvature directions, or adding new random ones - first only updating their model before including into proper step.
  \item Having the gradients, we can by the way use them for some first order optimization - like mentioned gradient descent in directions not included in second order model $(\tilde{g})$. It might be worth to explore more sophisticated hybrids of different order methods.
  \item Least squares linear regression could analogously provide 3rd order (or higher) local situation by additionally updating e.g. $\overline{\theta\theta\theta}$ and $\overline{g\theta\theta}$ averages - it might be worth considering  e.g. using 2nd order model in a few directions, additionally 3rd order in let say one dominant direction (e.g. as coefficient of its orthogonal polynomial), and 1st order gradient descent in the remaining. Their choice of dimensions could very adaptively.
  \item There are successful mechanisms for strengthening underrepresented coordinates, for example in AdaGrad or ADAM - they can be also applied in second order methods, what might be also worth exploring for example by increasing weights of such rare coordinates here.
\end{itemize}

\appendix The above article was written in 2019, this Appendix contains additional remarks (mainly from late 2022) for the proposed Hessian estimation like diagonal variants, Hessian symmetrization, corr=1 approximation, also simple implementations and low dimensional tests of the proposed OGR (online gradient regression) family of optimizers.

\subsection{Mean subtraction, subspace dimension and OGR variants}\label{amsub}
The central for discussed OGR (online gradient regression) is Hessian estimator, which can be seen as being built of (co)variance estimators, requiring to subtract $\bar{\theta},\bar{g}$ means (exponential moving here) - what is often missing in other approaches like ADAM (also $\sigma(\theta)$ in nominator), it offers improvements.

Denoting means with overline, as for covariance we get:
$$\hat{\theta}:=\theta-\bar{\theta}\qquad\qquad \hat{g}:=g-\bar{g}$$
$$\overline{\hat{g}\hat{\theta}^T}=\overline{(g-\bar{g})(\theta-\bar{\theta})^T}=
\overline{g\theta^T}-\overline{g}\,\overline{\theta}^T$$
$$\overline{\hat{\theta}\hat{\theta}^T}=\overline{(\theta-\bar{\theta})(\theta-\bar{\theta})^T}=
\overline{\theta\theta^T}-\overline{\theta}\,\overline{\theta}^T$$
\be H=\overline{\hat{g}\hat{\theta}^T}\ \overline{\hat{\theta}\hat{\theta}^T}^{-1}=\textrm{cov}(g,\theta)\ (\textrm{cov}(\theta,\theta))^{-1}\label{hesest}\ee
It seems beneficial to directly work on $\hat{\theta}, \hat{g}$ by first subtracting $\bar{\theta},\bar{g}$, then updating $\overline{\hat{\theta}\hat{\theta}^T}, \overline{\hat{g}\hat{\theta}^T}$ (later project to subspace), used in all further discussed simple implementations.
\subsubsection{Full Hessian (fOGR, cfOGR)} A basic approach is estimating full dimensional Hessian, with example simple implementation in Fig. \ref{beale}, practical rather only for low dimensional ($D$) problems. In this case we can just use the above mean subtraction, and update $D\times D$ matrices: $\overline{\hat{g}\hat{\theta}^T},\overline{\hat{\theta}\hat{\theta}^T}$ (or in later corr=1 approximation of cfOGR: $\overline{\hat{g}\hat{g}^T},\overline{\hat{\theta}\hat{\theta}^T}$).
\subsubsection{Evolving $d$ dimensional subspace (sOGR, lOGR, csOGR)} For high dimension we can use second order model in evolving $d<D$ dimensional local subspace, defined with evolving $d\times D$ projection matrix $V=(v_1,..,v_d)$ regularly shifted toward new gradients and (e.g. Gram-Schmidt) orthonormalized.

Its simple implementation is shown in Fig. \ref{dsOGR}, using $d$ number in name, e.g. s2OGR denotes $d=2$ dimensional subspace. For $s=1$ it becomes model along evolving line (as e.g. in ADAM), simplifying calculations (no eigendecompositions), so it might be worth treating separately: denoting lOGR $\equiv$ s1OGR.

As in this implementation, it is suggested to update $\bar{g},\bar{\theta}$ averages in full dimension $D$, subtract them from current $g,\theta$ then make projection to subspace, then update $d\times d$ matrices  $\overline{\hat{\theta}\hat{\theta}^T}$, $\overline{\hat{g}\hat{\theta}^T}$ using subtracted-then-projected $\hat{g}$, $\hat{\theta}$. This intuitively fixes to 0 center of subspace rotation - leading to smaller disturbance of $\overline{\hat{\theta}\hat{\theta}^T}$, $\overline{\hat{g}\hat{\theta}^T}$ averages with subspace rotation.

Initial basis could be chosen randomly, or better from the first gradients. Initial values of averages can be chosen as 0, with exception of $\overline{\hat{\theta}\hat{\theta}^T}$ which can be chosen e.g. as identity matrix times some small value. A more sophisticated initialization can improve behavior of the first steps. Estimated Hessian allows to rotate inside basis to diagonalize it, allowing for less frequent eigendecomposition.

\begin{figure}[t!]
    \centering
        \includegraphics[width=8.5cm]{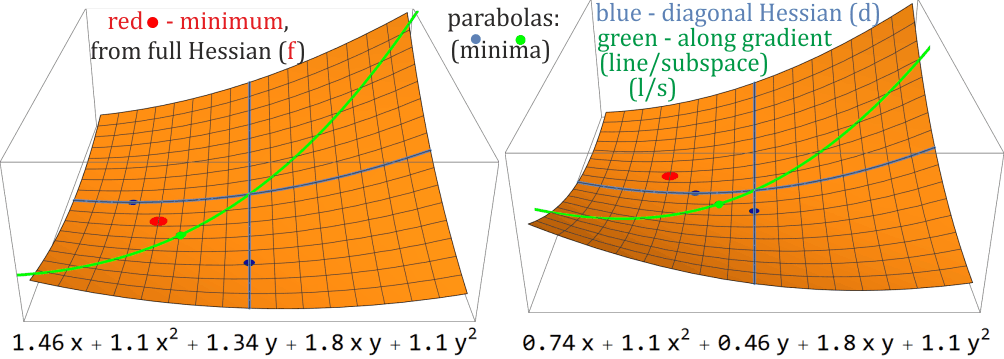}
        \caption{Examples of 2D situations for 2nd order polynomial - having full Hessian (estimated by fOGR, in subspaces by sOGR) and gradient, in one step we could reach the minimum (red dot). In contrast, using approximation of Hessian as diagonal (dOGR), we would use two blue parabolas, for each going (e.g. 1/div) toward its minimum (dot). Using line (subspace) approximation e.g. along gradient (s1OGR $\equiv$ lOGR), we focus on locally interesting direction. In example on the right, we can see that such directional approaches are imperfect - it would be beneficial to model Hessian in evolving subspace (sOGR) covering most of recent (local) gradient activity.  }
       \label{parab}
\end{figure}
\subsubsection{Choice of $d$ dimension, its potential evolution}
There is a difficult question of choosing considered dimension $d$, article \cite{tinysub} suggests most of evolution happens in $d\sim 10$ subspace we could use, maybe trying to tune, optimize it.

It would be beneficial to automatically choose $d$, maybe evolve it through optimization process. A natural evaluation of subspace activity is through (absolute values of) eigenvalues of modelled (sub)space Hessian. We can for example sort them $|\lambda_1|\geq |\lambda_2|\geq \ldots$, and choose the largest ones as above a threshold for minimal percentage of the largest eigenvalue, e.g. $|\lambda|>0.01 |\lambda_1|$.

This kind of evaluation would also allow to automatically evolve $d$ to adapt to local situation. We can e.g. remove direction from Hessian eigenbasis when its eigenvalue drops below the chosen threshold. Adding (increasing $d$) is more difficult: we first need to add a new vector to basis (e.g. momentum minus its projection on $V$), for a few steps just update averages for it, and then treat it as the remaining vectors. A condition for such basis increase could be e.g. last eigenvalue being above some larger threshold e.g. $|\lambda_d|>0.02 |\lambda_1|$.
\subsubsection{Diagonal Hessian approximation (dOGR,cdOGR)}
Calculations are much simpler if approximating Hessian as diagonal (dOGR variants) - zeroing nondiagonal dependencies, using separate parabola models. It allows to work on averages and Hessian as size $D$ vectors - update for $i=1,\ldots,D$ becomes:
$$\overline{\hat{g}\hat{\theta}^T}_{ii}' = (1-\beta) \overline{\hat{g}\hat{\theta}^T}_{ii} +\beta \hat{g}_i\hat{\theta}_i\qquad \overline{\hat{\theta}\hat{\theta}^T}'_{ii} = (1-\beta) \overline{\hat{\theta}\hat{\theta}^T}_{ii} +\beta \hat{\theta}_i\hat{\theta}_i$$
It allows to use very fast coordinate-wise multiplications of size $D$ vectors to model all $D$ parabolas, as e.g.  in Fig. \ref{dsOGR}, \ref{cdOGR}.

This simple variant builds independent parabola models for each of $D$ dimensions simultaneously in canonical basis, and use them to optimize step for each parameter separately - making it inexpensive and promising approach as least in low dimensions. Such independent parameter modelling might be also compatible with neural network.

The suggested 1D learning rate is $\lambda^{-1} = \overline{\hat{\theta}\hat{\theta}}/\overline{\hat{g}\hat{\theta}}$. The problem is that its denominator can be very small. As discussed, mathematically it is covariance, which can be written using correlation: $\overline{\hat{g}\hat{\theta}} = \textrm{corr}(g,\theta)\,\sigma(g)\sigma(\theta)$ for standard deviations: $\sigma(g)=\sqrt{\overline{\hat{g}\hat{g}}}$, $\sigma(\theta)=\sqrt{\overline{\hat{\theta}\hat{\theta}}}$. To prevent this covariance being low, as noticed in \cite{parabola}, we can use  $\textrm{corr}(g,\theta)=1$ approximation, giving safe learning rate $\lambda^{-1}=\sigma(\theta)/\sigma(g)$. Such simple "corr = 1" cdOGR variant is tested in further Fig. \ref{cdOGR} - getting better behavior thanks to safer choice (longer steps), also stuck less frequently - suggesting better generalization.
\subsubsection{Diagonal+subspace combinations (ldOGR/sdOGR)}
Low dimension evaluations in Fig. \ref{dsOGR}, \ref{cdOGR} suggest superiority of dOGR variants, however, sometimes low dimensional subspace models are misleading e.g. as in Fig. \ref{parab}, also neural network training often turns out focusing on a low dimensional subspace.

It suggests to combine (c)dOGR variant with sOGR: update both models simultaneously, and e.g. use a weighted average of their predicted step in the subspace (in perpendicular directions we can use dOGR step) - like in implementation in Fig. \ref{dsOGR}.

The weigh $w$ is additional hyperparameter, but usually we can choose $w\approx 1/2$, also we could  test which Hessian model has locally better prediction for difference of gradients ($\Delta g \approx H\cdot \Delta \theta$) - increasing weight of the more trusted one. While evaluation in Fig. \ref{dsOGR} has used common hyperparameters for 's' and 'd', it might be also worth optimize them separately.
\subsubsection{Online basis diagonalization} The subspace sOGR variants have freedom of internal rotation of the used basis as $d\times D$ orthonormal projection $V$. Having $d\times d$ Hessian model inside, we can diagonalize it $H=O^T D O$ and internally rotate $V\to OV$, making Hessian nearly diagonal inside such optimized basis. To reduce computational cost, instead of eigendecomposition, we can regularity perform step of e.g. QR method reducing non-diagonal coefficients.

The basic motivation is cost reduction, also removing  eigendecomposition from Hessian eigenvalue handling (e.g. div\&cut here). Also, we can use some additional methods to separately optimize behavior in such optimized (Hessian eigen-)directions, e.g. using evaluated trust levels, or higher order behavior.

\subsection{Symmetrized Hessian estimator}
The (\ref{hesest}) Hessian estimator is usually close to symmetric, but not exactly. We could manually symmetrize it e.g. using:
\be H = \frac{1}{4}\left(\overline{\hat{\theta}\hat{\theta}^T}^{-1}
\left(\overline{\hat{g}\hat{\theta}^T} + \overline{\hat{g}\hat{\theta}^T}^T\right) +\left(\overline{\hat{g}\hat{\theta}^T} + \overline{\hat{g}\hat{\theta}^T}^T\right)
\,\overline{\hat{\theta}\hat{\theta}^T}^{-1}\right)\label{hs}\ee
To find the proper formula, let us return to derivation from Section \ref{hesder} assuming $H=H^T$, also summing the necessity equations: $\partial_{H_{ij}}+\partial_{H_{ji}}=0$ leads to
$$ \overline{\hat{g}\hat{\theta}^T} + \overline{\hat{\theta}\hat{g}^T}=
H\,\overline{\hat{\theta}\hat{\theta}^T}+\overline{\hat{\theta}\hat{\theta}^T}\,H $$
To find $H$, let us decompose $\overline{\hat{\theta}\hat{\theta}^T}=ODO^T$ for $D=\textrm{diag}(\sigma^2)$, also denote $C=O^T
\left(\overline{\hat{g}\hat{\theta}^T} + \overline{\hat{g}\hat{\theta}^T}^T\right) O$ and $H'=O^T H O$. It allows to transform the above equation to
$$C=DH'+H'D\qquad\qquad C_{ij} = H'_{ij} (\sigma_i^2+\sigma_j^2)$$
which is symmetric, finally allowing to calculate $H$ as:
\be H=O\left[\left(O^T
\left(\overline{\hat{g}\hat{\theta}^T} + \overline{\hat{g}\hat{\theta}^T}^T\right) O\right)_{ij}/(\sigma_i^2+\sigma_j^2)  \right]_{ij}
O^T\label{hsf}\ee
with internal coordinate-wise division by $(\sigma_i^2+\sigma_j^2)$. Experimentally (\ref{hsf}) usually gives better optimization performance than (\ref{hs}), at cost of required eigendecomposition - which can be avoided if online diagonalizing (e.g. step of QR method)


\subsection{Hessian uncertainty}
While we know the exact positions $\theta$, unfortunately we rather do not know the exact gradients $g$, only estimate them from mini-batches, for randomized sample with uncertainty ("noise") of standard deviation $\sigma_n \propto 1/\sqrt{\textrm{mini-batch size}}$. While generally it can be direction-dependent, for simplicity let us assume here it is homogeneous noise - spherically symmetric with $\sigma_n$ standard deviation for each coordinate.

Now in time $t$,  $\bar{g}^t=\sum_{i\geq 0} \beta(1-\beta)^i g^{t-i}$ with weights summing to 1 hence it has the same $\sigma_n$ noise level. \\
For $\hat{g}^t=g^t-\bar{g}^t=\beta (\bar{g}^{t-1} - g)$ it grows to $\approx \sqrt{2} \beta \sigma_n$.

For Hessian estimator we need uncertainty of $\overline{\hat{g}\hat{\theta}^T}$. Approximating trajectory with the mean $\bar{\theta}$, standard deviation of $\overline{\hat{g}\hat{\theta}^T}_{ij}$ is approximately $\sqrt{2} \beta \sigma_n \bar{\theta}_j$.

\noindent Getting uncertainty of $H_{ij}$ as $\approx \sqrt{2} \beta \sigma_n \left(\textrm{diag}(\bar{\theta})\, \overline{\hat{\theta}\hat{\theta}^T}^{-1}\right)_{ij}$.

\noindent Of $\left(H^{-1}\right)_{ij}$ approximately
$\sqrt{2} \beta \sigma_n \left(H^{-1}\, \textrm{diag}(\bar{\theta})\ \overline{\hat{g}\hat{\theta}^T}^{-1}\right)_{ij}$.
\subsubsection{Division by uncertain value}
Let us look closer at 1D case (e.g. dOGR or in eigendirections), where estimated learning rate is $\overline{\hat{\theta}\hat{\theta}}/\overline{\hat{g}\hat{\theta}}$ along such single direction. The problem is that denominator is noisy here.

\begin{figure}[t!]
    \centering
        \includegraphics[width=8.5cm]{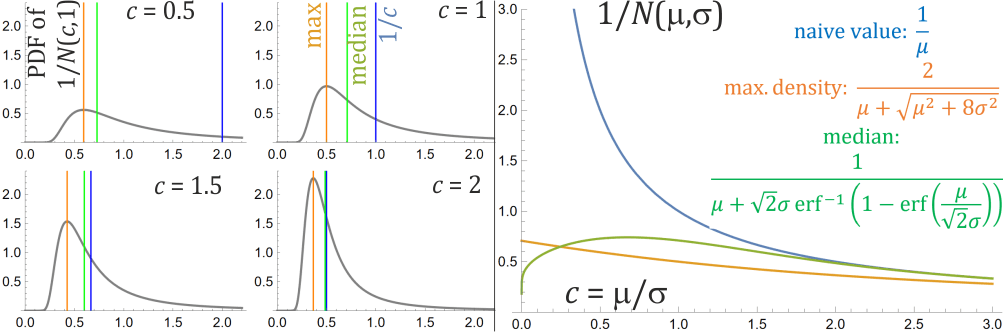}
        \caption{Regularization of noise in denominator (for $1/\overline{\hat{g}\hat{\theta}}$): examples of probability density functions of $1/N(c,1)$ inverted Gaussian distributions (generally $1/N(\mu,\sigma)$). Working on values, we would naively assume it is close to $1/c$, what is true only for large $c=\mu/\sigma$. Generally the expected value is infinite, there is shown density maximum (orange) and median (green) - we could use instead of $1/\mu$ as regularized inversion, maybe including absolute value $1/|N(\mu,\sigma)|$. Replacing value with a density model might be beneficial.}
       \label{rev}
\end{figure}

Figure \ref{rev} shows probability density functions for division by value from $N(\mu,\sigma)$ Gaussian distribution. If $\mu >> \sigma$ we can rather neglect the noise, but for small $\mu$ it becomes problematic, e.g. suggesting to increase mini-batch size.

As in this Figure, instead of just dividing by such value with uncertainty, as a regularization we can use some characteristic position of $1/N(\mu,\sigma)$ density, e.g. its maximum or median.

\subsection{Value calculation, augmented line search}
Beside gradient calculation, it seems worth to also regularly calculate values - on training set, or on validation/test set in early stopping - maybe also to help with generalization.

Another basic standard technique is line search - using e.g. 2nd order method to determine local direction, but not trusting it sufficiently to determine the distance - instead performing line search: calculating value in a few positions on this line and choosing position of the smallest value.

In contrast, as in shown simple implementations (in Fig. \ref{beale}, \ref{dsOGR}), the prosed approach is able to reasonably choose distance. However, being able to calculate values could help with speed of optimization process, for example using longer step (e.g. smaller \verb"div"), but then measuring value and e.g. finally using a smaller step, especially if the value has turned out larger (worse). There could be used smaller step of fixed e.g. 1/2 length, maybe more steps in such simple binary, line search.

It would be also worth to use the current estimated Hessian to augment such line search (one or a few steps), maybe also use the values to update the model e.g. through averages.

Assume such first step was suggested to be $\theta'=\theta-\Delta_\theta$, e.g. with SFN: $\Delta_\theta=\alpha |H|^{-1}m$ for $H$ current Hessian estimation and $m$ momentum. Line search would test $\theta'=\theta-a \Delta_\theta$ for various $a\in\mathbb{R}$. Taking absolute value as in SFN, we can assume $m^T\Delta_\theta>0$, the current model suggests behavior:
\be f_l(a):=f(\theta-a\Delta_\theta)\approx f(\theta)-a\, m^T\Delta_\theta + a^2\, |\Delta_\theta^T H \Delta_\theta|/2 \ee
Its inaccuracy has various sources: higher derivatives ($\Delta H\approx f'''\cdot \Delta\theta$), inaccuracy of momentum, and of gradient estimation. Additionally, we usually calculate values on some finite size (sub)set of training or test dataset, also making them noisy (contain inaccuracy).

Assuming we have calculated such (noisy) $f_l(0)$ and then $f_l(1)$, if the latter has turned out worse that the former, then we can e.g. try $f_l(1/2)$, and maybe continue such binary search.

One open question is making more educated guess than $a=1/2$, which intuitively seems perfect if $f_l(0)=f_l(1)$. However, if $f_l(1)$ turns out much worse than $f_l(0)$, then we could start with a smaller $a$.

The optimal choice of such $a$ would need to control both uncertainties, but an heuristic should be close, e.g. by comparing prediction with calculation - define their division as $b$:
 $$b=(f_l(0)-f_l(1))/(m^T\Delta_\theta - |\Delta_\theta^T H \Delta_\theta|/2)$$
Then $b=1$ means perfect prediction - if $\Delta_\theta$ was chosen as minimum of modeled parabola then we can stay there, also for $b>1$ meaning even better value drop than predicted.

\begin{figure}[t!]
    \centering
        \includegraphics[width=8.5cm]{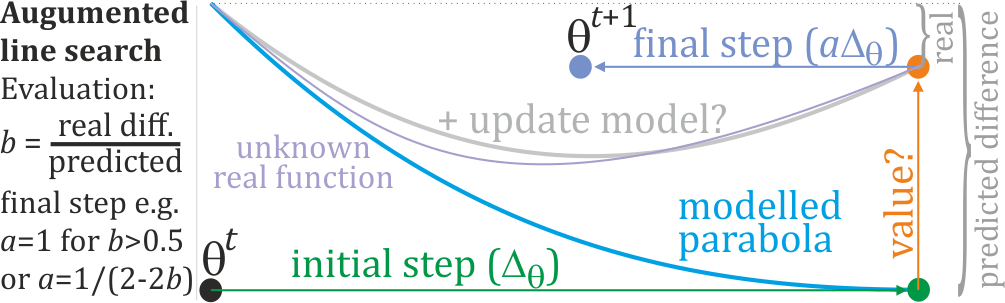}
        \caption{Augmented line search: if the used model suggests $\Delta_\theta$ step, we can calculate value in this position ($f(\theta-\Delta_\theta)$) and compare with model prediction - getting evaluation $b$. Then if this value was worse or not satisfactory, we can test/perform a shorter step in this line $f(\theta-a\Delta_\theta)$ e.g. for $a=1/(2-2b)$ when $b<1/2$, maybe also update the parabola model.}
       \label{line}
\end{figure}

However, for $b < 1$ we could try to improve. For $b<0$ we got worsening - should not make such step, at least try to improve with line search. For $b=0$ we get symmetric parabola - suggesting to use $a=1/2$, reduced for smaller $b$. Finally $a\to 0$ for $b\to -\infty$. Example of a simple formula agreeing with these constraints (to be optimized) is:
\be a = 1/(2-2b) \qquad\textrm{for e.g.}\qquad b\leq 1/2\ee
This way for $b>1/2$ we just accept the step, for lower we can e.g. use the above formula for the next step position, maybe further continuing e.g. using binary search or the same formula for modified $\Delta_\theta \to a \Delta_\theta$ in such augmented line search.

\subsubsection{Online value-based model update}
While above we have discussed the use of calculated values for augmented  line search, it is tempting to also use these values as additional local information to improve the model - e.g. the averages we use, especially $\bar{g}$ (and $m$ if separate), and $\overline{\bar{g}\bar{\theta}}$.

While gradient-based model update is at heart of discussed Hessian estimator, trying to analogously derive for values leads to more averages. We can instead try  approach: slightly modify the model to reduce its inaccuracy.

Denote current inaccuracy as $\Delta_f$:
$$\Delta_f=f(\theta')-f(\theta)+ m^T \Delta_\theta-\frac{1}{2} \Delta_\theta^T H\Delta_\theta$$
One reason of $\Delta_f\neq 0$ are higher order derivatives, others are various inaccuracies - we can try modifying $m$ and $H$ to reduce $\Delta_f$ inaccuracy, let say by some small $\xi$ percentage:
\begin{itemize}
  \item we can modify $m\to m-\xi\Delta_f\ \Delta_\theta/\|\Delta_\theta\|^2$,
  \item $H\to H+\Delta_H$ for $\Delta_H = 2\xi\Delta_f\ \Delta_\theta \Delta_\theta^T/\|\Delta_\theta\|^4$
\end{itemize}
Comparing with (\ref{hsf}), for symmetrized estimator the latter can be obtained by modifying/updating $\overline{\hat{g}\hat{\theta}^T}_s$:
$$\overline{\hat{g}\hat{\theta}^T}_s\to \overline{\hat{g}\hat{\theta}^T}_s+ O\left[\left(O^T \Delta_H
 O\right)_{ij}\cdot(\sigma_i^2+\sigma_j^2)  \right]_{ij} O^T $$
For dOGR we can modify $\overline{\hat{g}\hat{\theta}^T}$ coordinate-wise. The details are yet to be polished, but it could offer improvement over line search - additionally updating the model. Also, in case of nonsatisfying value, model update could suggest another promising step no longer being restricted to the $\theta-a\Delta_\theta$ line.

\subsection{Implicit OGR - separating gradient calculation position}
There are popular implicit e.g. Euler numerical scheme, in optimization e.g. as Nesterov Adaptive Optimizer~\cite{nesterov}, where we calculate gradient (maybe also value?) in a slightly shifted position (from $\theta$). Let us call the latter as $\theta_g$, which usually was $\theta_g=\theta$, but now we will separate them.

The main suggestion here is that the estimated Hessian is delayed: while being in $\theta^t$ position, the Hessian is estimated for current exponential moving average: $\bar{\theta}^t=\sum_{i\geq 0} (1-\beta)\beta^i \theta_g^{t-i}$ (assuming gradients calculated in $\theta_g^t$), and generally $\bar{\theta}^t \neq\theta^t$.

Hence the basic motivation for separation of $\theta_g$ from $\theta$ is trying to shift $\bar{\theta}$ closer to $\theta$, hopefully reducing the effect of higher derivatives. Here are two basic approaches for that:
\begin{itemize}
  \item After calculating the step $\theta'= \theta - \Delta_\theta$, we can calculate gradient and update the averages for slightly different position, e.g. for $\theta_g=\theta- \chi\Delta_\theta$ with $\chi$ hyperparameter
      slightly larger than 1.
  \item Optimize $\theta_g$ to shift new mean to the new position $\bar{\theta}=\theta'$:
  \be\theta' =\bar{\theta}'=(1-\beta)\bar{\theta}+\beta \theta_g \quad \Rightarrow \quad \theta_g = \frac{\theta-(1-\beta)\bar{\theta}}{\beta} \ee
\end{itemize}
There were performed some low dimensional tests for such simple modification, but not bringing essential improvements, what might change in higher dimensions.

\subsection{Online monitoring for hyperparameter update}
In practice, as benchmarks e.g. \cite{benchmark} show, it is often beneficial both to tune hyperparameters for specific tasks, also additionally schedule: evolve them through the optimization process (usually using fixed time evolution) - both can and should be considered for the discussed approach, e.g. reducing adaptation rates (especially $\Gamma$) in later phase of optimization.

However, it should be more beneficial to automatize it inside the optimization process, e.g. through some online monitoring trying to update - improve the current hyperparameters based on local situation, this way hopefully also properly choosing them in the first steps for a given specific task.

Let us discuss here some possibilities for such difficult but promising improvement directions.

\subsubsection{Trust level for gradient agreement}
Having current Hessian estimation, we can estimate change of gradients and compare it with the calculated difference - the better the agreement, the more trust  we can put in the model - like using longer steps (e.g. \verb"div" in implementations closer to 1).

The Hessian estimation is for $\bar{\theta}$ position, suggesting to use change of gradient from here: $g-\bar{g}\approx H (\theta-\bar{\theta})$.

Let us denote such trust evaluation as $T$, to handle noise it can e.g. undergo exponential moving average - below with adaptation rate $\nu$, to normalize it we can e.g. divide difference by sum of norms (like $l^2$ or $l^1$) - getting $T\in[0,1]$ local trust level evaluation (the lower the better):
\be T=(1-\nu)T+ \nu\, \frac{\|H(\theta-\bar{\theta})-(g-\bar{g})\|}{\|H(\theta-\bar{\theta})\|+\|g-\bar{g}\|} \label{trustl}\ee
The big question is how to translate such trust level into hyperparameters, mostly in the step size, e.g. something like \verb"div"$=1+T$ in suggested implementation, or some more sophisticated formula, up to applying a neural network here.
\subsubsection{Coordinate-wise trust levels e.g. for dOGR}
In dOGR variant (or sOGR with online Hessian diagonalization) we can treat all coordinates (or Hessian eigendirections) independently, also choose their trust levels separately. For this purpose we can just calculate (\ref{trustl})-like formula for 1D values with coordinate-wise vector functions, and somehow use the trust levels e.g. to choose separate \verb"div" step sizes.

\subsubsection{Updating separate models for various parameters}
A basic universal approach is maintaining multiple models, here especially averages for various adaptation rates $\beta$, and continuously monitoring their accuracies - e.g. choosing locally the best model, or weighting between suggestions of separate models - the better recent accuracy, the higher the weight.

For example using above trust level (\ref{trustl}) to evaluate multiple Hessian models for various $\beta$, and using $1-T$ as weights divided by their sums, and choosing step as such weighted average of steps proposed by individual models.

Such set of hyperparameters could be also modified, e.g. through parameter evolution, removing those having the worst evaluations, adding new ones, maybe also using some crossover between parameter sets as in evolutionary algorithms.

Analogous trust level application could be choice of weight in dsOGR - updating two models, the locally more accurate one should be assigned larger weight.

\subsubsection{Direct optimization of adaptation rates like $\beta$}
We could try to analogously optimize adaptation rate $\beta$ parameter, which controls speed of change of estimated Hessian ($H\to H'$) - we can try to optimize it to make new Hessian in better agreement with the observed change of gradients ($g-\bar{g} \approx H'(\theta-\bar{\theta}))$.

Let us denote next values with prime, previous without:
\be\overline{g}'=(1-\beta) \overline{g} +\beta g\qquad \qquad \hat{g}=g-\bar{g}\label{summ}\ee
$$\overline{\theta}'=(1-\beta) \overline{\theta} +\beta \theta\qquad\qquad \hat{\theta}=\theta-\bar{\theta}  $$
$$\overline{\hat{g}\hat{\theta}^T}' = (1-\beta) \overline{\hat{g}\hat{\theta}^T} +\beta \hat{g}\hat{\theta}^T\qquad \overline{\hat{\theta}\hat{\theta}^T}' = (1-\beta) \overline{\hat{\theta}\hat{\theta}^T} +\beta \hat{\theta}\hat{\theta}^T$$
Let us use the last to look at update of Hessian estimator, using (\ref{invm}), $H=\overline{\hat{g}\hat{\theta}^T}\, \overline{\hat{\theta}\hat{\theta}^T}^{-1}$ and defining $\bar{\beta}=\beta/(1-\beta)$:
$$H'=\overline{\hat{g}\hat{\theta}^T}'\, \overline{\hat{\theta}\hat{\theta}^T}'^{-1}=\left(\overline{\hat{g}\hat{\theta}^T} +\bar{\beta} \hat{g}\hat{\theta}^T\right) \left( \overline{\hat{\theta}\hat{\theta}^T} +\bar{\beta} \hat{\theta}\hat{\theta}^T\right)^{-1}\approx $$
$$\approx \left(\overline{\hat{g}\hat{\theta}^T} +\bar{\beta} \hat{g}\hat{\theta}^T\right)\,
\overline{\hat{\theta}\hat{\theta}^T}^{-1}\,
 \left(I -\bar{\beta} \hat{\theta}\hat{\theta}^T \, \overline{\hat{\theta}\hat{\theta}^T}^{-1}\right) $$
\be H'\approx  H + \bar{\beta}\left(
\hat{g}\hat{\theta}^T -
H\, \hat{\theta}\hat{\theta}^T
\right)\overline{\hat{\theta}\hat{\theta}^T}^{-1}
 +O(\bar{\beta}^2)\ee
From the other side, Hessian defines change of gradient with position:  $\Delta g \approx  H \Delta \theta$. We can for example use differences from the means here:
$$g-\bar{g} \approx \left(H + \bar{\beta}\left(
\hat{g}\hat{\theta}^T -
H\, \hat{\theta}\hat{\theta}^T
\right)\overline{\hat{\theta}\hat{\theta}^T}^{-1}\right)(\theta - \bar{\theta})$$
\be g-\bar{g} - H(\theta - \bar{\theta}) \approx \bar{\beta}\left(
\hat{g}\hat{\theta}^T -
H\, \hat{\theta}\hat{\theta}^T
\right)\overline{\hat{\theta}\hat{\theta}^T}^{-1}(\theta - \bar{\theta})\label{updt}\ee
Calculating vectors on the left and right, their good/bad agreement suggests that we can/cannot trust the model e.g. decreasing/increasing $\beta$. Natural evaluation of vector agreement is scalar product, cosine - the open question are details of using it here, to evolve $\beta$ or maybe modify it for single steps. Also step size and change of Hessian might be included.

\begin{figure*}[t!]
    \centering
        \includegraphics[width=18cm]{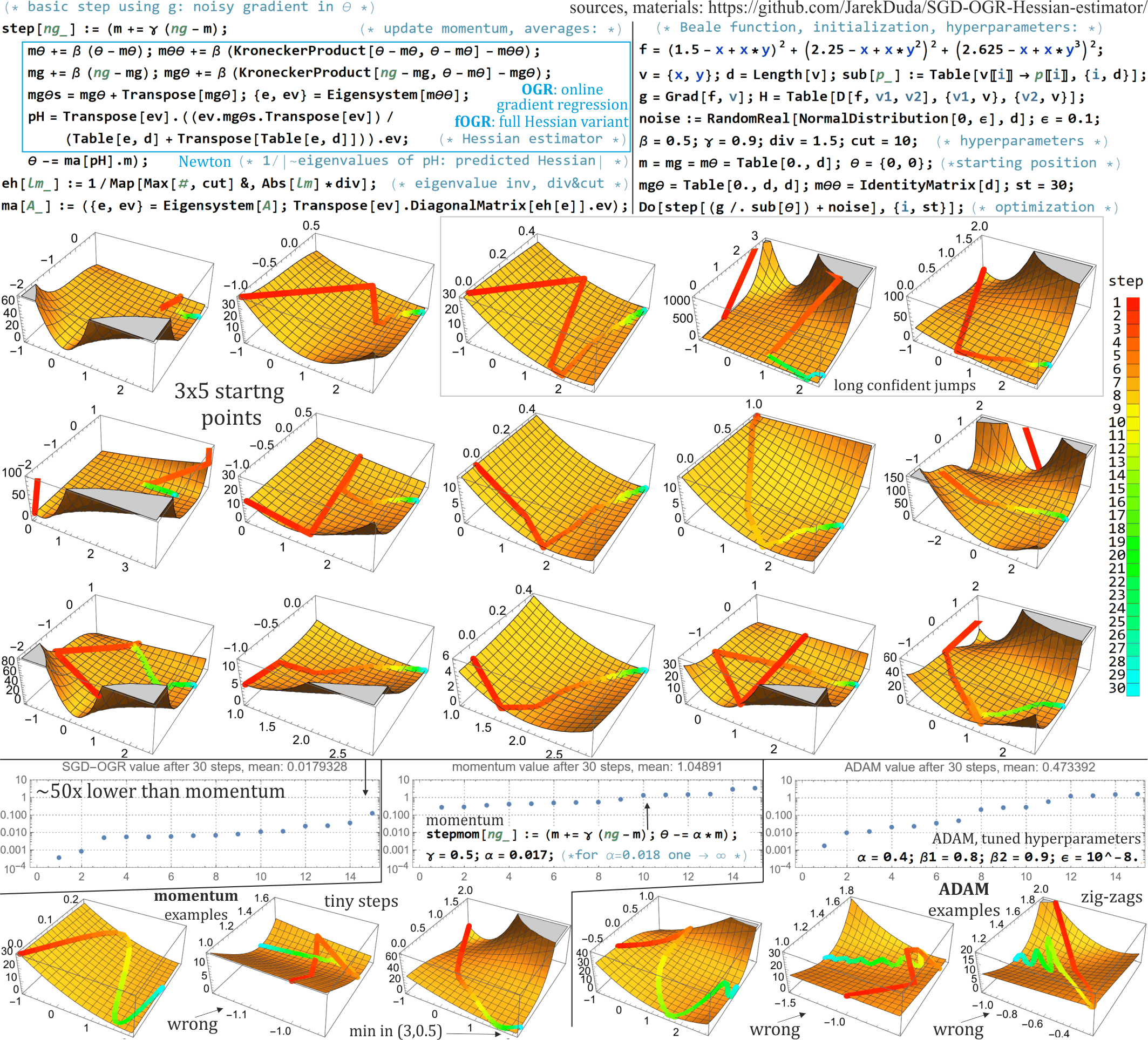}
        \caption{Beale function optimization using shown fOGR variant, with various starting points: $\{-2,-1,0,1,2\}\times \{-1,0,1\}$. We can see usually in $\approx 10$ steps (yellow) it starts approaching the minimum in $(3,0.5)$. It uses separate learning rates for momentum ($\gamma$) and symmetrized (\ref{hsf}) Hessian estimator ($\beta$). Suggested $1/\lambda$ learning rates undergo absolute value as in SFN, then are divided by "div", then bounded by "1/cut". For neutral network training it should evolve in locally interesting e.g. 10 dimensional subspace~\cite{tinysub}, chosen e.g. by online PCA~\cite{OPCA} of gradients or proposed here exploration of new directions. At the bottom there are 3 examples of momentum (slow) and ADAM (zig-zags) - for starting positions as marked top right 3 for OGR (long confident jumps).}
       \label{beale}
\end{figure*}

\subsection{Adding regularizer to Hessian estimator}\label{hreg}
In Section \ref{hesder} there was derived central here MSE Hessian estimator from 4 (e.g. exponential moving) averages. Let us  add here regularizer preferring use of low Hessian coefficients, by adding the $r(H)$ penalty to the previously optimized:
$$\arg\min_{H,p}\ \sum_{i,t} w^t \left(g^t_i-\sum_k H_{ik}(\theta^t_k-p_k)\right)^2+r(H) $$
Repeating the previous derivation with this additional term $r(H)$ adds its derivation, using $\hat{g}=g-\bar{g}$ we get:
%
\be \overline{\hat{g}\hat{\theta}^T}=
H\,\overline{\hat{\theta}\hat{\theta}^T}+
\partial_H r(H)\label{der}\ee
for $\partial_H\equiv [\partial_{H_{ij}}]_{ij}$ matrix of derivatives.
\subsubsection{Hessian $l^2$ regularizer} a natural choice:
$$r(H)=\frac{1}{2}\eta \sum_{ij} (H_{ij})^2$$
gives $\partial_H r(H) = \eta H$, getting simple natural formula for $l^2$ regularized Hessian estimator:
\be H=\overline{\hat{g}\hat{\theta}^T}
\left(\overline{\hat{\theta}\hat{\theta}^T}+\eta I \right)^{-1}\ee
While it might be valuable for Hessian estimation, Newton step uses inverted $H^{-1}g$, suggesting e.g. to add $\eta I$ to nominator instead - let us now try to formalize it.
\subsubsection{Inverted Hessian $l^2$ regularizer} replacing $H\to H^{-1}$: $r(H)=\frac{1}{2}\eta \sum_{ij} ((H^{-1})_{ij})^2$ and using standard formula:
\be (M+\Delta)^{-1}= M^{-1}-M^{-1}\Delta M^{-1}+O(\|\Delta\|^2)\label{invm}\ee
leads to $\partial_{H} r(H) = -\eta (H^{-1})^T H^{-1} (H^{-1})^T$.

While Hessian is symmetric, this MSE estimation is not exactly (further we discuss symmetrization), but we can use approximation $\partial_{H} r(H) \approx  -\eta H^{-3}$ making (\ref{der}):
$$\overline{\hat{g}\hat{\theta}^T}\approx
H\,\overline{\hat{\theta}\hat{\theta}^T}-\eta H^{-3}$$
$$H^{-1}\approx
\left(\overline{\hat{\theta}\hat{\theta}^T}-\eta H^{-4}\right) \overline{\hat{g}\hat{\theta}^T}^{-1}$$
Which could be used with the right hand side $H^{-1}$ as some approximation, e.g. $H^{-1}$ from the previous step. However, it is difficulty to avoid infinities in $\overline{\hat{g}\hat{\theta}^T}^{-1}$, suggesting to shift $\eta$ to denominator - what can be approximately done using (\ref{invm}):
\be H^{-1}\approx\overline{\hat{\theta}\hat{\theta}^T}\left(\,\overline{\hat{g}\hat{\theta}^T}+\eta H^{-3}\right)^{-1} \label{reginv}\ee
Still to avoid singularities, we would need to replace  $\overline{\hat{g}\hat{\theta}^T}$ with a nonnegative matrix, e.g. replacing eigenvalues with their absolute values as in SFN.

\begin{figure*}[t!]
    \centering
        \includegraphics[width=18cm]{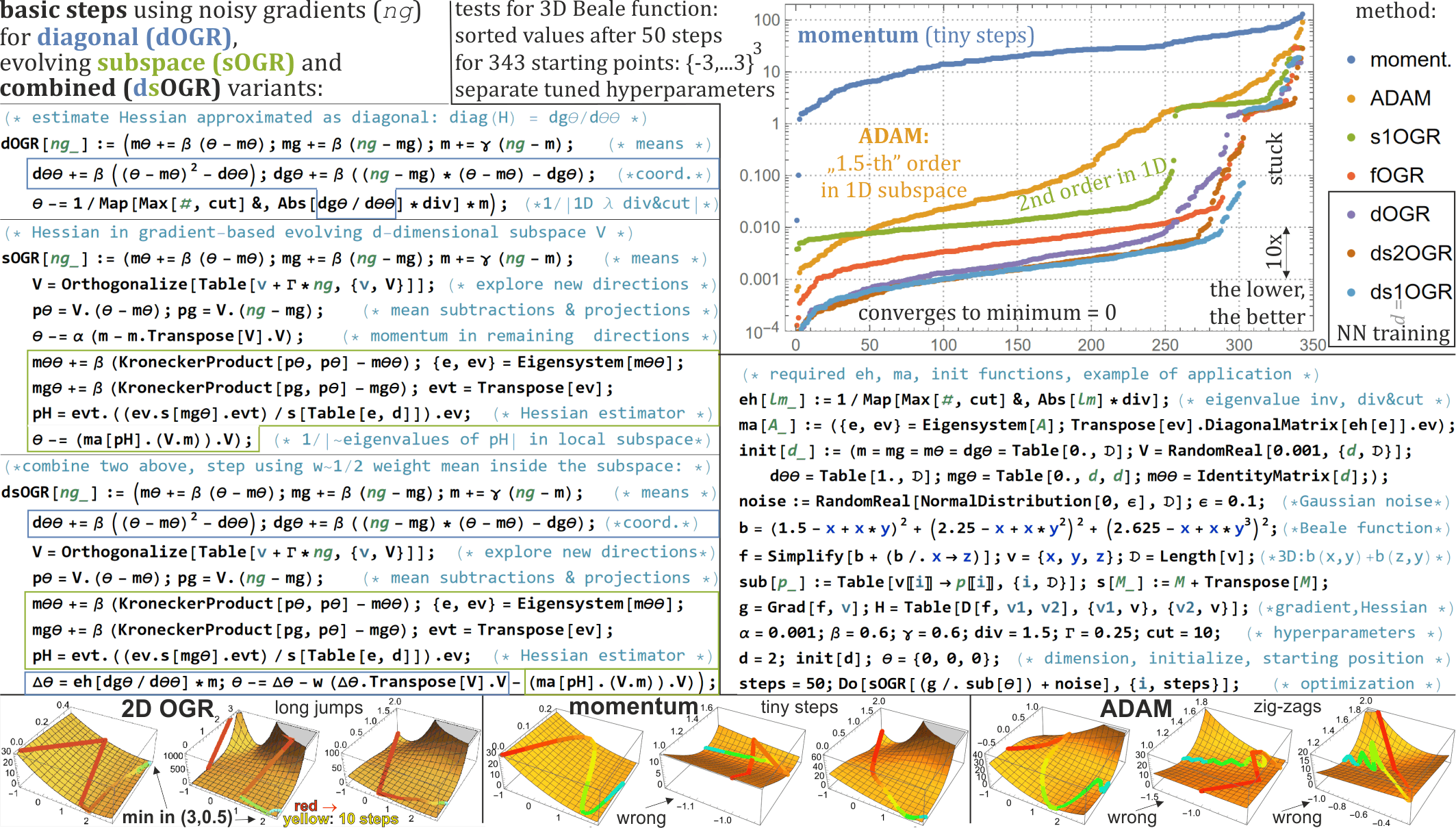}
        \caption{Simple implementations of further OGR variants (dsOGR.nb file in github.com/JarekDuda/SGD-OGR-Hessian-estimator/) intended for neural network training (d and s, but not f) - to be used in high dimension $D$. The diagonal version dOGR models separate parabolas for each direction, what requires updating 4 (m=mg) or here 5 size $D$ vectors, using coordinate-wise vector multiplications/divisions. The subspace sOGR variants model Hessian in (written in name) $d<D$ dimensional locally interesting subspace - regularly updated by adding $\Gamma$ times gradient to basis vectors and applying Gram-Schmidt orthogonalization. There is also shown dsOGR variant updating both models, and inside the subspace taking weighted ($w$) average of two predicted steps. Shown evaluation has used written 3D Beale function starting from $\{-3,\ldots 3\}^3$ size 343 set of points, there are shown sorted 343 values after 50 steps, with individual hyperparameter tuning for geometric mean evaluation - the lower, the better. Simple dOGR gave excellent results, which can be slightly improved augmenting with subspace models - could be beneficial for NN training, where evolution mostly happens in nearly constant low dimensional subspace~\cite{tinysub}. }
       \label{dsOGR}
\end{figure*}

\subsection{Practical bound for $H^{-1}$} Let us discuss here extension of cdOGR "corr = 1" bound to higher dimension. The Newton step uses $H^{-1}g$, which can go to infinity - it is crucial to prevent such behavior, somehow clipping the step, bounding $H^{-1}$ which eigenvalues work as learning rates in corresponding eigendirections. It has $\overline{\hat{g}\hat{\theta}^T}$ covariance matrix in denominator:
$$H^{-1}=\overline{\hat{\theta}\hat{\theta}^T}\ \overline{\hat{g}\hat{\theta}^T}^{-1}=
\textrm{cov}(\theta,\theta)\ (\textrm{cov}(g,\theta))^{-1}$$
We can perform PCA (principle component analysis): find $$\textrm{cov}(\theta,\theta)=O_\theta \textrm{diag}(\sigma_\theta^2) O_\theta^T\qquad  \textrm{cov}(g,g)=O_g \textrm{diag}(\sigma_g^2) O_g^T$$
with $\sigma$ as vectors of standard deviations, $\sigma^2$ is coordinate-wise. Then transform $\tilde{\theta}=O_\theta^T \hat{\theta} O_\theta$ from approximately normal distribution with standard deviations given by coordinates of $\sigma_\theta$, and analogously for $\tilde{g}=O_g^T \hat{g} O_g$ with $\sigma_g$ standard deviations. This way Hessian estimator becomes:
\be H=O_g\, \overline{\tilde{g}O_g^T O_\theta \tilde{\theta}^T}\ \textrm{diag}(\sigma_\theta^{-2}) O_\theta^T \ee
The difficult part is $O_g^T O_\theta$ difference of PCA basis for gradients and positions. However, changes of positions are made accordingly to gradients, suggesting to approximate $O_\theta \approx O_g$ as single $O$ from gradients, which can be used in the discussed online basis diagonalization to reduce cost (matrix $\to$ diagonal as vector).
\be H^{-1}_{\textrm{safe}} \approx O\, \textrm{diag}(\sigma_\theta/\sigma_g)\, O^T\ee
This is maximal correlation approximation, as suggested for 1D in \cite{parabola}, now extended to multiple directions. It essentially bounds $H^{-1}$ allowing for $\sigma_\theta/\sigma_g$ safe choice of learning rates.

This approximation neglects non-diagonal dependencies, but it is done in locally optimized basis from PCA of gradients. Some its improvements could be obtained e.g. using SVD, Canonical Correlation Analysis.

\subsubsection{Direct corr=I multidimensional approximation} While there are various ways to take  corr=1 approximation to higher dimensions, brief numerical exploration suggests to just directly use
\be \textrm{cov}(g,\theta) \approx \sqrt{\textrm{var}(g)} \sqrt{\textrm{var}(\theta)}\quad \textrm {for covariance matrices} \ee
with square root acting on matrix - its eigenvalues. Then for such $\textrm{cov}(g,\theta)$ approximation we can use symmetrized Hessian estimator (\ref{hsf}).

Thanks to avoiding the division by 0 problem, the improvement turned out excellent - available in GitHub, shown in Figure \ref{cOGR}. Low dimension numerical experiments suggests this approximation allows to have nearly compete trust in parabola model: div = $1/\eta \approx 1$. However, sometimes steps are too large - suggesting to use "cut" mechanism, or adding small value to denominator, or maybe some line search.

\begin{figure}[t!]
    \centering
        \includegraphics[width=8.5cm]{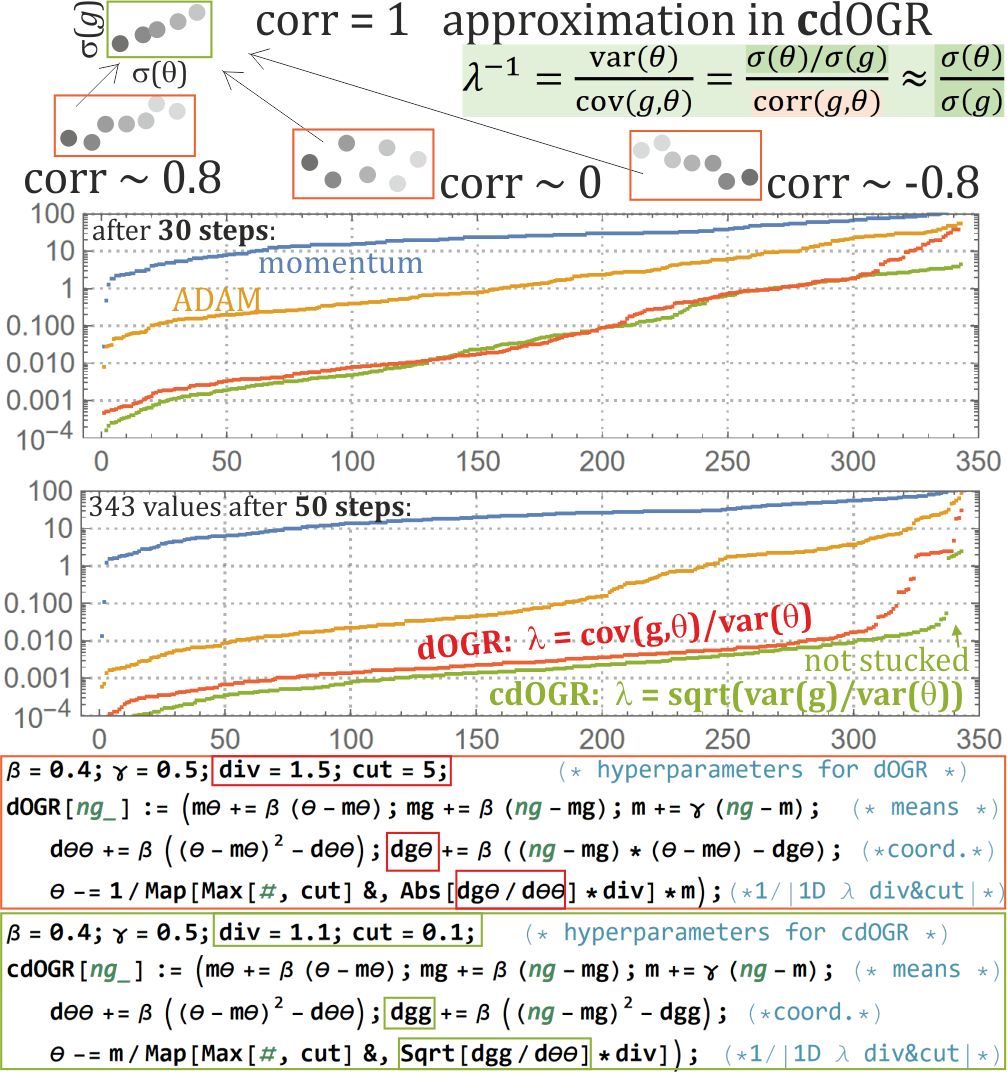}
        \caption{Improved "correlation=1" diagonal dOGR variant (separate parabola models in all directions) using $\textrm{corr}(g,\theta)=1$ approximation (hence $\lambda\approx \sigma(g) /\sigma(\theta)$) to avoid division by small values, what has allowed for more confident steps (smaller $\textrm{div}\approx 1$ and cut).
        Using $\gamma=\beta$ we can remove one average (m=mg), however, separating them with larger $\gamma$ (faster adaptation of momentum in Newton step) usually gives slightly better performance.  Interestingly, it turned out to much less frequently stuck in false solutions, bringing hope for being better at generalization. }
       \label{cdOGR}
\end{figure}

\subsection{Higher order methods}
While full 2nd order method is rather impractical in high dimensions: requiring $D^2$ size Hessian, full higher order methods would require even higher powers of $D$  for tensor of derivatives - making them even less practical.

However, as e.g. dOGR here, using separate models for each coordinate brings excellent performance, and analogously could be extended to include higher derivatives e.g. as d3OGR for separate 3rd order model in each direction.

Alternatively, we could use the evolving $d<<D$ dimensional subspace approach (line for $d=1$):  $d^2$ size Hessian inside, $d^3$ size tensor for 3rd power, and so on. Intermediate approach could use Hessian to diagonalize subspace basis, and for its coordinates (Hessian eigendirections) use separate 3rd order models to improve choice of step sizes.

To find analogous 1D higher order formulas, let us consider linear regression of gradients $g$ using family of functions e.g. $h_i(x)=x^i$ for polynomials, and overline as the average (exponential moving here) for varying $(\theta,g)\in\mathbb{R}^2$ pairs:
$$\textrm{optimization problem:}\qquad\arg\min_{(a_i)} \overline{\left(g-\sum_{i=0}^\textrm{order} a_i h_i(\theta)\right)^2 }$$
$$\forall_j\quad 0=-\delta_{a_j}=\overline{h_j(\theta)\left(g-\sum_i h_i(\theta)\right)}$$
$$\forall_j\quad \overline{h_j(\theta)g}=\sum_ia_i \overline{ h_i(\theta) h_j(\theta)}$$
\be \textrm{coef. vector:}\quad (a_i)_i =\left(\left[\overline{ h_i(\theta) h_j(\theta)} \right]_{ij}\right)^{-1} (\overline{h_j(\theta)g})_j \label{ho} \ee
For order 2 polynomials it leads to the previous formulas. For order 3 we would need updating additional 3 averages: $\overline{\theta^3},\overline{\theta^4},\overline{g\theta^2}$, and the formulas are much more complicated due to matrix inversion in (\ref{ho}), even if using mean subtraction.

In practice it seems worth to treat higher order information as additional helping with choice of step sizes - e.g. vector of div for cdOGR predicted e.g. with some machine learning methods from trust levels, but also e.g. $\overline{(g-\bar{g})^3}$, $\overline{(g-\bar{g})^4}$.
\subsection{Adding a neural network optimizing use of local information}
While there were discussed various approaches for improvements, finding the details seems very difficult and problem/architecture dependant. It suggests to maybe train a small neural network (or a different machine learning technique) trying to optimize behavior based on all gathered local information. Especially for diagonal 'd' variants - updating averages $\overline{\hat{g}^p}, \overline{\hat{\theta}^q}$ for various powers $p,q$ (not necessarily natural), maybe also $\overline{\hat{g}^p\hat{\theta}^q}$, maybe also such averages for various adaptation rates, maybe also trust levels (evaluations of $\Delta g \approx H \Delta \theta$ agreement), numbers of step (scheduling), maybe some additional e.g. architecture information, and trying to predict a perfect learning rates e.g. for each canonical direction.

To train such neural network (or other model), we can start with trajectories for a standard method like ADAM: gathering such local information along trajectory, and also finding perfect step lengths: from line search in each direction. Then train this small neural network based on such pairs: (local information, perfect step), and use it to choose steps.

As such trained model would change the trajectory, we can later gather new training data for its trajectories, retrain and so on. It might be also worth to consider more sophisticated e.g. LSTM neural networks here - trying to continuously adapt to the current specific problem during optimization.

\subsection{Simple implementation and additional remarks}
There were prepared simple Mathematica implementations for OGR family\footnote{Sources, materials:  github.com/JarekDuda/SGD-OGR-Hessian-estimator/}, together with results for popular 2/3D Beale function (\url{https://en.wikipedia.org/wiki/Test_functions_for_optimization}) shown in Fig. \ref{beale}, \ref{dsOGR}. They use mean subtraction (\ref{summ}) (before subspace projection). For Newton $H^{-1}m$ step they use coordinate-wise dOGR or symmetric Hessian estimator (\ref{hsf}), and separate learning rates for $m$ momentum ($\gamma$, fastest adaptation), $H$ Hessian estimator ($\beta$, slower), and subspace update ($\Gamma$, the slowest). For subspace update it just adds gradient times $\Gamma$ to all basis vectors (might be worth normalizing before), and apply Gram-Schmidt orthonormalization.

For saddle repulsion they use absolute values of Hessian eigenvalues as SFN, also divide suggested by Hessian learning rates by \verb"div", and bound them by \verb"1/cut". Further optimization might e.g. use separate \verb"div", \verb"cut" for positive and negative eigenvalues - reduced learning rates for negative as suggested e.g. by simulations in \cite{hess}. There could be optimized more sophisticated than such div\&cut behaviors, e.g. with some nonlinearity like based on discussed above regularized estimators. It seems worth to modify  hyperparameters during optimization (scheduling), e.g. reduce \verb"div" toward 1 in late stage to speedup convergence in safe situations, reduce $\Gamma$ for slower subspace rotations. Other mentioned improvement directions are to be considered in the future.

While these f/s implementations use two eigendecompositions, its computational cost in e.g. 10 dimensions (suggested in \cite{tinysub}) is rather negligible in comparison to batch gradient calculation. In much higher dimension, as discussed such Hessian estimation can focus on locally interesting e.g. 10 dimensional subspace (and tiny learning rate in perpendicular directions): obtained e.g. by online PCA of gradients or discussed here cheaper exploration of new dimensions. It can be combined with more or less frequent online Hessian diagonalization e.g. regular QR method steps for savings in eigendecompostion.

The Figures also contain simple tests to compare with other approaches, like momentum - which often escapes to infinity: requiring small learning rate, slow evolution. In contrast popular ADAM has longer jumps, but we can zig-zags as it locally thinks one-dimensionally, also the formulas are heuristic (e.g. missing mean subtraction) - in OGR replaced by real 2nd order method: formally derived from linear regression of gradients, also simultaneously optimizing in multiple dimensions - allowing for long confident jumps.

\begin{figure*}[t!]
    \centering
        \includegraphics[width=18cm]{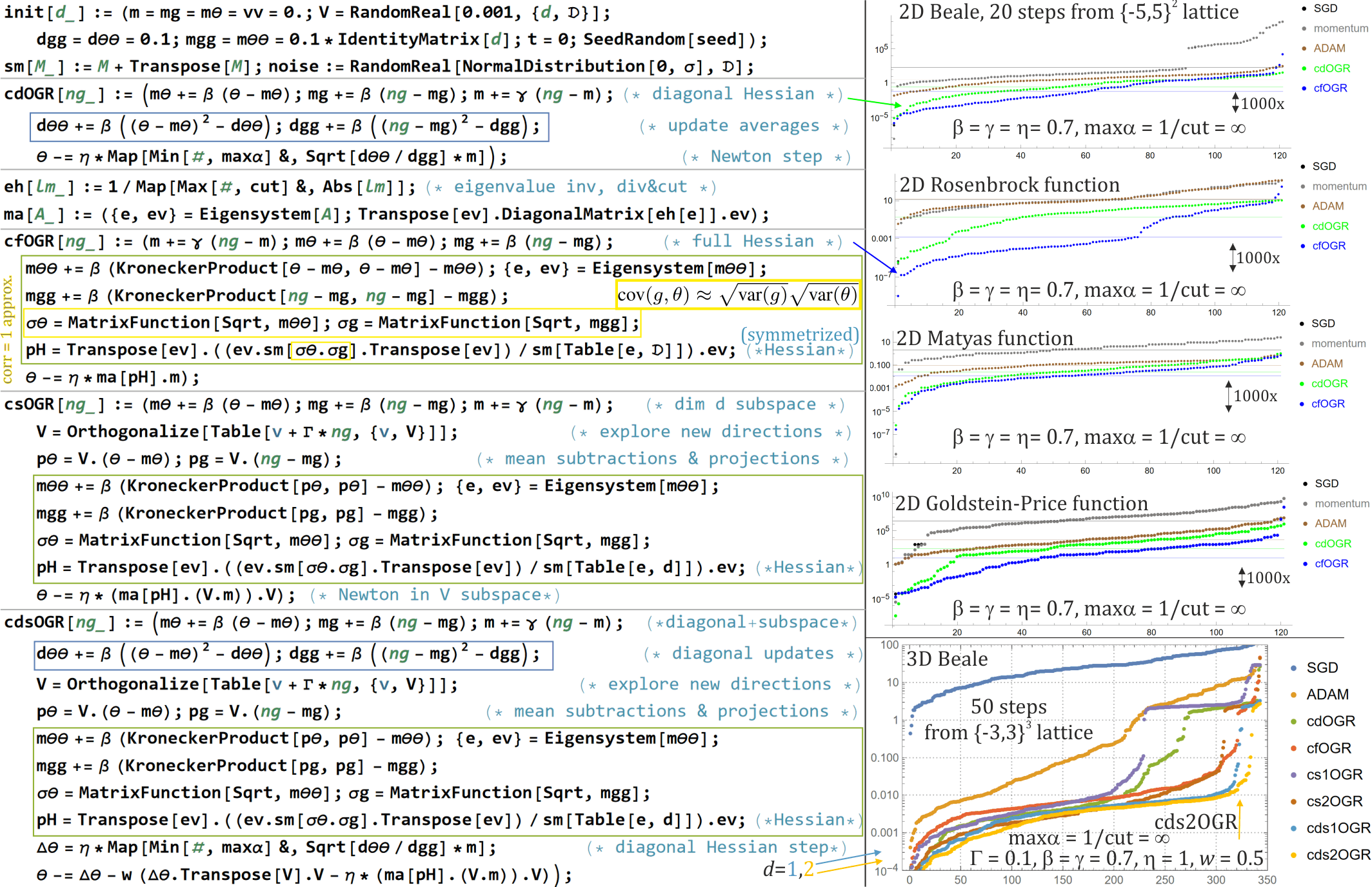}
        \caption{Various 'c' variants: using corr=1 approximation: $\textrm{cov}(g,\theta) \approx \sqrt{\textrm{var}(g)} \sqrt{\textrm{var}(\theta)}$, previously in standard 1D case in cdOGR (separate for each canonical direction). Now such approximation is also used for multidimensional Hessian e.g. in cfOGR, csOGR - with square roots as matrix functions (acting on eigenvalues), approximating $\textrm{cov}(g,\theta)$ as $\textrm{var}(g)$, $\textrm{var}(\theta)$ eigenvectors being perfectly correlated. There are also shown evaluations (GitHub) as previously (sorted values starting in lattice after fixed numbers of step) - in 2D for popular test functions (from \url{https://en.wikipedia.org/wiki/Test_functions_for_optimization}) using written trivial hyperparameters, and for 3D Beale function as previously in Fig. \ref{dsOGR}. We can see large benefits for multi-dimensional Hessian model of cfOGR in 2D, often $\sim$1000x smaller values than ADAM in 20 steps. For 3D Beale there is good convergence for nearly all starting points (generalization?) especially when combining diagonal and ($d=1$ or $2$ dimensional) subspace Hessian models in cdsOGR variants, then weighting their predicted steps with trivial $w=1/2$ weight here. Surprisingly, $\eta=1/$div$=1$ worked perfectly here, what means complete trust in such parabola models with corr=1 approximation.  }
       \label{cOGR}
\end{figure*}

Figure \ref{cdOGR} contains later approach for corr = 1 approximation for separate coordinates, later extended to multidimensional approximation and combined with subspace variants in Figure \ref{cOGR} - providing essential improvements.\\

Some additional remarks:
\begin{itemize}
 \item At least in low dimensions dOGR, cdOGR give excellent performance, could be improved combining with sOGR for additional optimization in locally active directions, maybe also e.g. with trust level evaluation, higher order information e.g. $\overline{(g-\bar{g})^3}$, $\overline{(g-\bar{g})^4}$ and other mentioned techniques.
 \item The symmetrized Hessian estimator (\ref{hsf}) experimentally gives significant improvement for optimizer.
 \item The corr=1 approximation of "c" variants essentially improves behavior by avoiding division by low values. However, in theory avoiding this approximation might allow for even better behavior - e.g. using line search or even some small neural network optimizing use of various local information, trying to predict perfect step sizes.
 \item It is worth to continuously update the Hessian model, preferably with small batches, maybe with some uncertainty control, in let say $d=10$ dimensional locally interesting subspace (e.g. \cite{tinysub}). In directions perpendicular to considered subspace, we can still use momentum method with tiny learning rate ($\alpha$ in Fig. \ref{dsOGR}), or dOGR in dsOGR.
 \item Details of Newton step for estimated Hessian is an open question worth optimization, e.g. including regularizer, choose separate behaviors for positive and negative eigendirections~(\cite{hess}), trust evaluation, etc.
  \item It seems worth to continuously monitor agreement of our model with calculated gradients and maybe values, e.g. trying to update step lengths (up to separately for each direction in dOGR), adaptation rates, maybe use values to modify averages (especially $\overline{\hat{g}\hat{\theta}}$), or locally change behavior e.g. to some (augmented) line search.
  \item It is be worth to use multiple adaptation rates (maybe e.g. given by function of a single parameter): the fastest for gradient used in Newton step (momentum), slower for 4 averages in Hessian estimator, and the slowest for the $(v_i)$ vectors defining locally interesting subspace ($\Gamma$ in Fig. \ref{dsOGR}). It might be also worth to evolve adaptation rates through some scheduling, or even use multiple simultaneously and  combine their predictions by some weighting.
  \item It might be worth separating positions for gradient/value calculation, as discussed for implicit OGR.
\item For generalization we would like to find wide minimum, avoiding narrow ones. Such focus could be enforced by smoothing the function, for example by adding Laplacian times some constant - it could be done here inside the considered low dimensional subspace, e.g. using trace of estimated Hessian as approximated Laplacian. Experimentally some variants are better in avoiding fake local minima (especially cdsOGR) - might have generalization.
\end{itemize}

\bibliographystyle{IEEEtran}
\bibliography{cites}
\end{document}